\documentclass[letterpaper, 10 pt, journal, twoside]{ieeeconf} 

\IEEEoverridecommandlockouts                              

\usepackage[paper=letterpaper,margin=0.75in]{geometry}

\usepackage{graphicx} 
\usepackage{epstopdf}
\DeclareGraphicsExtensions{.eps}

\usepackage{multirow}
\usepackage{makecell}
\usepackage{tabularx}
\usepackage{algorithmic}
\usepackage{algorithm}
\usepackage{amsmath} 
\usepackage{amssymb}
\usepackage{amsxtra}
\usepackage{url} 
\usepackage{color} 
\usepackage{bm}
\usepackage{placeins}
\usepackage[caption=false]{subfig}
\usepackage{xspace}
\usepackage{rotating}
\usepackage{siunitx}
\usepackage{textcomp}%

\newcommand{\myvector}[1]{\bm{#1}}
\newcommand{\myvec}[1]{\myvector{#1}}

\newcommand{\R}[1]{\mathbb{R}^{#1}}

\newcommand{\algorithmicinput}{\textbf{Input:}}
\newcommand{\algorithmicoutput}{\textbf{Output:}}

\newcommand{\INPUT}{\item[\algorithmicinput]}
\newcommand{\OUTPUT}{\item[\algorithmicoutput]}

\newcommand{\dist}[1]{\SI{#1}{\meter}}

\newcommand{\excise}[1]{}

\newcommand{\revised}[2]{\textcolor{black}{#2}}

\newif\ifremark
\long\def\remark#1{
  \ifremark%
  \begingroup%
  \dimen0=\textwidth
  \advance\dimen0 by -1in%
  \setbox0=\hbox{\parbox[b]{\dimen0}{\protect\em #1}}
  \dimen1=\ht0\advance\dimen1 by 2pt%
  \dimen2=\dp0\advance\dimen2 by 2pt%
  \vskip 0.25pt%
  \hbox to \textwidth{%
    \vrule height\dimen1 width 3pt depth\dimen2%
    \hss\copy0\hss%
    \vrule height\dimen1 width 3pt depth\dimen2%
  }%
  \endgroup%
  \fi}

\newcommand{\ui}[1]{a}

\newcommand{\alg}{RL-RRT}

\newcommand{\revisedV}[1]{\textcolor{black}{#1}}
\usepackage{hyperref}

\title{
\LARGE \bf
RL-RRT: Kinodynamic Motion Planning via Learning Reachability Estimators from RL Policies
}

\author{Hao-Tien Lewis Chiang$^{1,2}$, Jasmine Hsu$^{1}$, Marek Fiser$^{1}$, Lydia Tapia$^{2}$, Aleksandra Faust$^{*1}$
\thanks{$^{1}$Google AI, Mountain View, CA 94043, USA {\tt\scriptsize {lewispro,hellojas,mfiser,sfaust}@google.com}}%
\thanks{$^{2}$University of New Mexico, Albuquerque, NM 87101, USA {\tt\scriptsize {tapia}@cs.unm.edu}}%
\thanks{$^*$ Corresponding author}
\thanks{Accepted as regular paper to Robotics and Automation Letters in June 2019.}
}

\begin{document}

\maketitle
\thispagestyle{empty}
\pagestyle{empty}

\begin{abstract}
This paper addresses two challenges facing sampling-based kinodynamic motion planning: a way to identify good candidate states for 
local transitions and the subsequent computationally intractable steering between these candidate states.  Through the combination of sampling-based planning, a Rapidly Exploring Randomized Tree (RRT) and an efficient kinodynamic motion planner through machine learning, we propose an efficient solution to long-range planning for kinodynamic motion planning.
First, we use deep reinforcement learning to learn an obstacle-avoiding policy that maps a robot's sensor observations to actions, which is used as a local planner during planning and as a controller during execution.
Second, we train a reachability estimator in a supervised manner, which predicts the RL policy's time to reach a state in the presence of obstacles. 
Lastly, we introduce RL-RRT 
that uses the RL policy as a local planner, and the reachability estimator as the distance function to bias tree-growth towards promising regions. We evaluate our method on three kinodynamic systems, including  physical robot experiments.
Results across all three robots tested indicate that RL-RRT  outperforms  state of the art  kinodynamic planners in efficiency, and also provides a shorter path finish time than a steering function free method. 
The learned local planner policy and accompanying  reachability estimator demonstrate transferability to the previously unseen experimental environments, making RL-RRT fast because the expensive computations are replaced with simple neural network inference. Video: \href{https://youtu.be/dDMVMTOI8KY}{https://youtu.be/dDMVMTOI8KY}

\end{abstract}


\section{INTRODUCTION}
\label{sec:intro}

Consider
motion planning for robots such as UAVs \cite{liu2017planning}, autonomous ships \cite{chiang2018colreg}, and spacecrafts \cite{richards2002spacecraft}. The planning solution needs to satisfy two criteria. First, the solution path must be feasible, meaning that the path must be collision-free and satisfy kinodynamic constraints, e.g. velocity and acceleration bounds even in the presence of sensor noise.
Second, the path needs to be efficient, i.e. near optimal with respect to objectives such as time to reach the goal. For example, a motion plan for a car-like robot should avoid obstacles,  reach the goal promptly, not make impossibly sharp turns, and maintain enough clearance to compensate for sensor noise.


\begin{figure}[t]
	\centering
	\subfloat[\scriptsize RL-RRT and SST in Map 1 (46.1 x 49.5 m) ]{\includegraphics[width=0.35\textwidth,height=4.0cm,keepaspectratio=true]{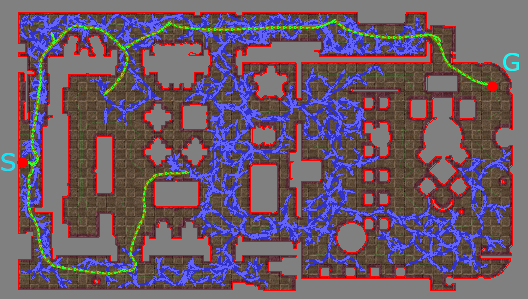}\label{fig:coverPic}}
	
	\subfloat[\scriptsize The Fetch robot]{\includegraphics[height=4cm,keepaspectratio=true]{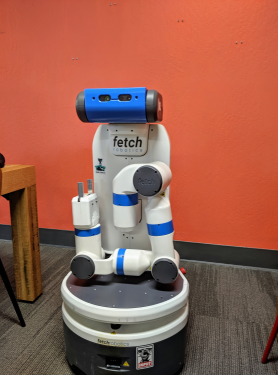}\label{fig:fetchPic}}
	\subfloat[\scriptsize Trjectory execution of Fetch in Map 2 (46.1 x 49.5 m) ]{\includegraphics[width=0.33\textwidth,height=4cm,keepaspectratio=true]{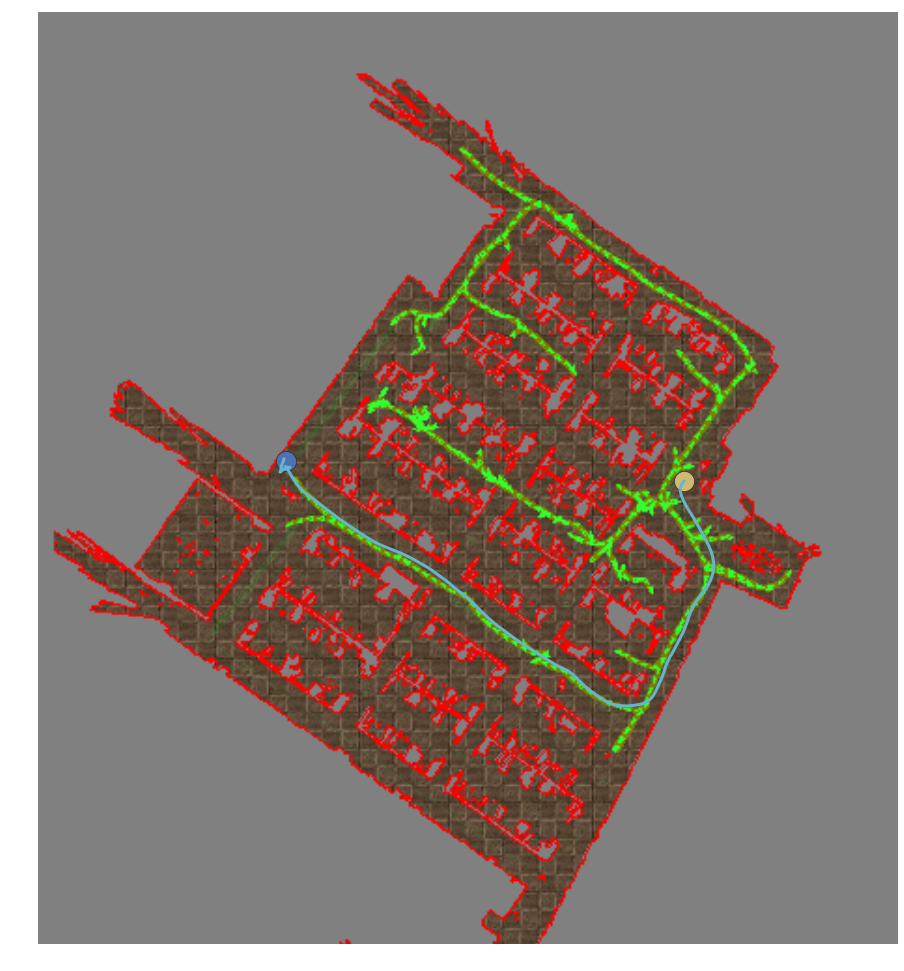}\label{fig:physicalPath}}
	\caption{
\revised{}{(a) Example trees constructed with RL-RRT (yellow) and SST  \cite{li2016asymptotically} (blue) for a kinodynamic car navigating from start (S) to goal (G).
    (b) The Fetch robot. (c) RL-RRT (green) and the real-world trajectory executed (cyan) from the start (green dot) towards the goal (blue dot) in Map 2. Map 2 is a SLAM map of an actual office building.}
	}
	\label{}
\end{figure}

\revised{E.2}{Current state of the art kinodynamic motion planners 
search the robot's feasible state space by building a tree data structure of possible robot motions rooted at the robot's current state. The methods iteratively use a local planner to attempt to grow the tree until the goal is reached. 
While some tree-based methods grow the tree by randomly propagating actions, others guide the tree growth to focus state space expansion thus requiring the local planner to be a steering function, a control policy that guides a robot to a specific goal in obstacle-free space, while satisfying the kinodynamic constraints. For example, consider a car-like robot needing to translate a small distance to the left, a motion resembling parallel parking. This motion plan is difficult, even in the absence of obstacles, and requires a steering function to steer the car to the goal. Computing the steering function requires solving an optimal control problem, and is generally NP-Hard  \cite{wolfslag2018rrt}. To date, only very limited robot dynamics such as linear  \cite{webb2013kinodynamic} and differential drive \cite{palmieri2015distance} systems have optimal steering functions.
}

\revised{E.2}{Besides the existence of steering functions, there are two additional difficulties facing efficient tree-based kinodynamic motion planning. First, tree-based methods that use steering functions require identifying the state in the tree from which to grow.  This requires a function that compare the distance between states and return those that are expected to be easily solved by the steering function.  
An effective distance function for kinodynamic planning is the Time To Reach (TTR) between states using an optimal steering function \cite{palmieri2015distance}.
TTR, however, is often expensive to compute as it involves numerically integrating the steering function \cite{palmieri2015distance}. 
Second, neither the steering functions nor the related TTR are informed by sensors, and, as a result, do not account for potential obstacles. For example, if a goal is occluded by a wall, the steering function is not able to see the wall due to the lack of sensory input, and TTR would return a value as if an agent could go through the wall.}

Recently, deep Reinforcement Learning (RL) emerged as a promising near optimal steering function for kinodynamic systems \cite{layek2017deep}.
\revised{}{In addition, deep RL algorithms can learn policies that map noisy lidar or camera observations directly to robot actions, thus enabling obstacle avoidance while navigating between states for differential drive robots \cite{autorl,unstuck-dinesh}.}
\revised{}{With the recent development of AutoRL \cite{autorl}, which uses evolutionary algorithms to largely eliminate the need to hand-tune hyper-parameters, network structure and reward functions.} This combination offers the promise of deep RL being employed for local planning, i.e., providing both steering function and obstacle avoidance.
However, RL policies often lack long-term planning capabilities \cite{mnih2015human} and get trapped in environments with complex obstacles \cite{prm-rl}.

To address \revised{E.2}{the lack of available steering functions, good distance functions for aiding tree growth, and obstacle-awareness} facing kinodynamic motion planning, we propose \alg, which combines RL and sampling-based planning.
It works in three steps. 
First, we learn an obstacle-avoiding point-to-point (P2P) policy with AutoRL. 
\revised{E.2}{This is a mapless, goal-conditioned policy that maps sensor readings to control. These P2P policies generalize to new environments without re-training \cite{autorl}. 
Second, we train a supervised obstacle-aware reachability estimator that predicts the time it takes the P2P policy to guide the robot from a start to goal state in presence of obstacles, using local observations such as lidar. The key insight is that the AutoRL policy and the estimator implicitly learn the topology of the obstacles. This allows reasonably accurate estimates of time to reach in new environments.}
Lastly, \revised{E.2}{presented with a motion planning problem in a new envrionment,} in a RRT setting, we use the RL policy as a local planner and the reachability estimator as the distance function.   The combination of these two learning solutions offers two primary advantages.  
First, by using RL policies as an obstacle avoiding local planner, RL-RRT can be applied to a variety of kinodynamic systems without optimal steering functions. 
Second, by using the obstacle-aware reachability estimator, RL-RRT can prune out randomly sampled states that are un-reachable from the tree, e.g., the policy is expected to be unsuccessful, and identify nodes with small TTR to the sampled state. \revised{E.2}{In the example of a car in front of a wall, the RL policy will go around the wall, and the estimator will predict that the time to reach will be longer because of the wall.}

We evaluate RL-RRT in two environments with three kinodynamic robots.
Results indicate that AutoRL policies are effective obstacle-avoiding local planners. 
The obstacle-aware reachability estimators, one for each robot, are 74-80\% accurate in identifying if a goal state is reachable.
Compared to a state of the art steering function free method, 
SST \cite{li2016asymptotically}, RL-RRT is up to 2.3 times more likely to identify a path within a fixed time budget and the identified path is up to 4.5 times shorter.
RL-RRT typically identifies dynamically-feasible paths in very few iterations -- 11 in this case -- thanks to intelligent node selection and the obstacle-avoiding local planner (Figure \ref{fig:coverPic}).
The enclosed video demonstrates RL-RRT tree construction and trajectory execution on a physical differential drive robot.

\section{RELATED WORK}
\label{sec:related-work}

Steering function-based kinodynamic planners, such as kinodynamic RRT* \cite{webb2013kinodynamic} and D-FMT \cite{schmerling2015optimal}
rely on a steering function to ``pull'' the tree to achieve rapid exploration \cite{phillips2004guided} and a proper distance function  \cite{webb2013kinodynamic,palmieri2015distance,wolfslag2018rrt}. 
\revised{}{RL-RRT uses AutoRL \cite{autorl} to learn steering functions, thus bypassing the challenging two-point boundary value problem.}

\revised{}{Steering function free-based approaches, such as EST \cite{phillips2004guided} and SST \cite{li2016asymptotically}, propagate random actions from a selected node. 
These methods can be applied to a variety of robot dynamics, although they tend to ``wander'' \cite{allen2016real}, thus they can take a long time to identify a solution.}  

Recent research has offered several solutions for P2P obstacle-avoidance policies on a differential drive robot from raw sensory input, including learning from demonstration \cite{Pfeiffer2017FromPT}, curriculum learning \cite{successor-features}, and reinforcement learning \cite{virtualtai2017,autorl}. 
Other research offers hierarchical solutions to  navigation, where the RL agent executes a path identified by another planner, e.g., from a grid \cite{unstuck-dinesh}, PRMs \cite{prm-rl, francis2019long}, or manually selected waypoints \cite{ddqn-topological}.
However, none of those methods are designed for kinodynamic robots, leading to failures at milestones due to dynamic constraints \cite{francis2019long}.


Designing an effective distance function for sampling-based kinodynamnic motion planning is challenging \cite{palmieri2015distance}.
The commonly used Euclidean and weighted Euclidean distance for configuration space planning is inefficient as kinodynamic robot states have limited reachability \cite{li2011learning}. 
The minimum TTR between states is a highly effective distance function \cite{palmieri2015distance,wolfslag2018rrt} but is often too computationally-expensive to be used as a distance function \cite{palmieri2015distance}.
While learned TTR of a near-optimal differential drive steering function \cite{palmieri2015distance} can overcome the computational complexity, this approach still requires a near-optimal steering function.
Indirect optimal control has also been used to generate training samples composed of minimum TTR and optimal control actions along trajectories \cite{wolfslag2018rrt}.
However, this approach currently only works for low dimensional systems such as inverted pendulum and does not handle limited action bounds. 
\revised{}{Our approach addresses} these challenges by bypassing the necessity of a near-optimal steering function via RL. 
Unlike previous methods, we also take into account obstacle avoidance, which can significantly change the minimum TTR. 


\section{METHODS}
\label{sec:methods}

\revised{3.7}{RL-RRT is a kinodynamic motion planner that learns local planner and distance function w.r.t the individual robot dynamics.
It has three main steps.}
First, we learn an obstacle-avoiding point to point policy with AutoRL \cite{autorl}.
Next, since the RL policy avoids obstacles, we can use the policy to generate obstacle-aware reachability training samples by repeatedly rolling out the learned policy. 
An obstacle-aware reachability estimator is trained to predict the time to reach between two robot states in the presence of obstacles. \revised{E.2}{Policy and estimator training is done once per robot in training environments.} 
Third, during planning, RL-RRT creates dynamically-feasible motion plans using the RL policy as the local planner and the reachablity estimator as the distance function. \revised{E.2}{Note, that the training and planning simulators require simulated depth measurements (e.g. lidar) around the robot, which can be thought of as analogous to motion planning with information about clearance.}


\subsection{AutoRL Local Planner}
\label{sec:methodAutoRl}

We train a RL agent to perform a P2P task that avoids obstacles. 
The output of the training is a policy that is used as a local planner, an execution policy, and a data generation source for the obstacle-aware reachability estimator. 
\revisedV{Using one RL policy for both local planning and path execution is inspired by \cite{francis2019long}. This allows the planner to account for potential noise during path execution.}
\revisedV{To train a policy robust against noise, we model the RL policy is a solution for a continuous state, continuous action, partially observable Markov decision process (POMDP) given as a tuple $(\Omega, S, A, D, R, \gamma, O)$ of observations, state, actions, dynamics, reward, scalar discount, $\gamma \in (0,1) $, and observation probability.} 
The observations are the last three measurements of the noisy robot lidar \revisedV{ and potentially noisy relative goal position and robot velocity.} 
\revisedV{We define states as the true robot configuration and its derivative.}
A black-box robot dynamics simulator, which maps states-action pairs to states, is an input to the RL training environment. 
\revisedV{Another black-box simulator maps the robot state to noisy lidar observations w.r.t. obstacles.}
The goal is to train the agent to reach a goal state, $G$, within radius, $d_{G}.$
\revised{2.1}{Note that AutoRL identifies a policy that maps noisy sensor and state observations to action. 
We explore simulated lidar measurement noise in this work and left state estimation and process noise to future work.}
\revised{}{AutoRL training is required only once for a given robot.}

AutoRL \cite{autorl} over DDPG \cite{ddpg}, used for learning the RL agent policy, takes as input: observations, actions, dynamics, goal definition, $(G, r)$, and a parametrized reward, $R:O \times \theta_r \rightarrow \R{},$. 
\revised{}{The agent is trained to maximize the probability of reaching the goal without collision.} 
This is achieved by using evolutionary algorithms over populations of agents to find a dense reward that maximizes successful goal reaching. Each generation of agents is trained with a new reward, selected based on the previous experience. At the end, the fittest agent that performs P2P tasks best, is selected as the P2P policy. In this work, all three agents use the same observations, goal definitions, and neural network architectures, but differ in the robot dynamics and reward features used. 

As an example, we explain the training of the Asteroid robot here (details of the robot are in the Appendix). Details for the Differential Drive and Car robot can be found in \cite{autorl} and \cite{francis2019long}.
\revisedV{The observation is a vector of $3N_\text{beams}$ noisy lidar returns concatenated with the relative planar position of the goal, the robot velocity and orientation ($3N_\text{beams}+5$ dimensional vector).}
\revisedV{The state is the planar position, velocity and orientation of the robot.}
The action is the amount of forward thrust and turn rate.
The parameterized reward includes
\begin{equation}
R_{\myvec{\theta_{r_\text{DD}}}} = 
\myvec{\theta}^T [
r_\text{goal}
r_\text{goalDist} \,
r_\text{collision} \,
r_\text{clearance} \,
r_\text{speed} \,
r_\text{step} \,
r_\text{disp} \,
],
\nonumber
\label{eq:asteroid_reward}
\end{equation}
where
$r_\text{goal}$ is 1 when the agent reaches the goal and 0 otherwise,
$r_\text{goalDist}$ is the negative Euclidean distance to the goal,
$r_\text{collision}$ is \revisedV{-1} when the agent collides with obstacles and 0 otherwise,
$r_\text{clearance}$ is the distance to the closest obstacle,
$r_\text{speed}$ is the agent speed when the clearane is below 0.25m,
$r_\text{step}$ is a constant penalty step with value 1, and
$r_\text{disp}$ is sum of displacement between the current and positions 3, 6 and 9 steps before.
\revisedV{$\myvec{\theta}$ is the weight vector tuned by AutoRL.}

\subsection{Obstacle-Aware Reachablity Estimator}
\label{sec:methodReachability}


We further improve upon work in \cite{palmieri2015distance} by learning the TTR of an obstacle-avoiding P2P RL policy learned in Section \ref{sec:methodAutoRl}.
Our obstacle-aware reachability estimator provides the following benefits:
1) It does not need an engineered near-optimal steering function for each robot dynamics. This allows TTR learning for robot systems without near-optimal steering functions.
2) Due to the presence of obstacles, the minimum TTR between states is a function of both robot dynamics and obstacles. 
Since RL policies can also learn to avoid obstacles, the obstacle-aware reachability estimator can provide additional benefit over TTR estimators that consider only obstacle dynamics such as \cite{palmieri2015distance}. 



\subsubsection{Training data collection}
\begin{algorithm}[tb]
	\caption{Training data collection} 
	\label{alg:trainingData}
\begin{algorithmic}[1]
\footnotesize
\INPUT  $\myvec{\pi}(\myvec{o})$: Obstacle avoiding P2P RL policy, $N_\text{episode}$: Number of episodes, $\Delta t$: Time step size, $T_\text{horizon}$: Reachability horizon
\OUTPUT $\text{trainingData} = (\myvec{o}_1, y_1), (\myvec{o}_2, y_2), \cdots, (\myvec{o}_N, y_N)$.
\FOR {$i=1,\cdots N_\text{episode}$ }
    \STATE $\myvec{s}, \myvec{g}$ = sampleStartAndGoal()
    \STATE elapsedTime = 0
    \WHILE{isDone is False}
        \STATE elapsedTime += $\Delta t$
        \STATE $\myvec{o}$ = makeObservation()
        \STATE executePolicy($\myvec{\pi}(\myvec{o})$, $\Delta t$)
        \STATE obsHistory.append($\myvec{o}$)
        \STATE \revised{2.3}{$c$}, isDone = \revised{2.3}{getTTRCost}(elapsedTime, $T_\text{horizon}$)
        \STATE \revised{2.3}{costHistory}.append(\revised{2.3}{$c$})
    \ENDWHILE
    \STATE \revised{2.3}{cfc} = \revised{2.3}{computeCumulativeFutureCost}(\revised{2.3}{costHistory})
    \FOR {j=0, len(obsHistory)}
        \STATE trainingData.append(($\myvec{o}=$ obsHistory[j], $y=$ \revised{2.3}{cfc}[j]))
    \ENDFOR
    \STATE obsHistory.clear(); \revised{2.3}{costHistory}.clear()
\ENDFOR
\RETURN trainingData
\end{algorithmic}
\end{algorithm}

Algorithm \ref{alg:trainingData} summarizes the training data collection. 
First, for each episode, we initialize the robot with randomly chosen start and goal states (Alg. \ref{alg:trainingData} line 2).
Next, we execute the policy until the episode terminates (lines 4-11) \revised{3.1}{due to reaching the goal, collision, or reaching a time horizon $T_\text{horizon}$.}
During execution, we record the robot observation at each time step (line 8) and
 compute and record the TTR \revised{2.3}{cost} (lines 9-10).
The TTR \revised{2.3}{cost} is set to $\Delta t$ at every time step.
\revised{3.1}{To classify whether the robot can reach the goal, we use a simple heuristic that penalizes trajectories that do not reach the goal.
If the robot is in collision or the time horizon is reached (elapsedTime equals to $T_\text{horizon}$), the TTR cost of that time step is set to $\Delta t + T_\text{horizon}$, and the episode is terminated immediately by setting isDone to true.
}

After an episode terminates, \revised{2.3}{we compute the cumulative future TTR cost for all states along the trajectory, i.e., remaining cost-to-go to the end of the trajectory (line 12)}.
The observation and cumulative future cost of each time step form a training sample and is recorded (line 14).
The process repeats for $N_\text{episode}=1000$ episodes.
\revised{3.1}{We designed the TTR cost heuristic such that if the robot reaches the goal, the cumulative future cost of each state along the trajectory is the TTR between that state and the goal.}
Conversely, if the robot failed to reached the goal due to collision or the episode reaches time horizon, all \revised{3.1}{cumulative future cost} along the trajectory will be larger than $T_\text{horizon}$.
\revised{3.1}{By employing a common machine learning technique that uses a regressor and a threshold value as a classifier \cite{goodfellow2016deep}, 
we can quickly classify whether a goal state can be reached during planning.}

\subsubsection{Reachability Estimator Network}
We train the obstacle-aware reachability estimator network with the training data collected above.
The network input is the robot observation $\myvec{o}$ and the output is the estimated TTR.
We use a simple three-layer fully-connected network with [500, 200, 100] hidden neurons with each a dropout probability of 0.5.
We use the L2 loss between estimated TTR and the V-value label from the training data.

\subsection{\alg}
Alg. \ref{alg:rlrrt} describes \alg.
While the standard RRT algorithm was utilized,  modifications were made to efficiently utilize the obstacle-aware reachability estimator and the obstacle-avoiding RL local planner.


\begin{algorithm}[tb]
\begin{algorithmic}[1]
\footnotesize
\INPUT  $\myvec{\pi}(\myvec{o})$: Obstacle avoiding P2P RL policy, $\Delta t_\text{tree}$: Tree extension time step size, $\Delta t$: policy time step size, $T_\text{horizon}$: Reachability horizon, $P_\text{goalBias}$: Goal bias, $\myvec{x}_\text{root}$: Current robot state, $k_c$: Number of candidate nodes
\OUTPUT $\mathcal{P}$: Motion plan.

\STATE iteration = 0
\STATE $\mathcal{T}$.add(makeNode($\myvec{x}_\text{root}$, None))
\WHILE {termination condition not met}
    \STATE iteration += 1
    \STATE goodXrndFound = False
    \WHILE {not goodXrndFound}
        \STATE $\myvec{x}_\text{rnd}$ = sampleCollisionFreeStateSpace($P_\text{goalBias}$)
        \STATE candidateNodes = findNearestNodesEu($\mathcal{T}$, $\myvec{x}_\text{rnd}$, $k_c$)
        \STATE $n_\text{nearest}$ = findNearestNode(candidateNodes, $\myvec{x}_\text{rnd}$)
        \STATE TTR = getAvgTTR($n_\text{nearest}$, $\myvec{x}_\text{rnd}$)
        \IF {TTR $<$ TTR$_\text{threshold}$ or rnd $> P_\text{prune}$}
            \STATE goodXrndFound = True
        \ENDIF
    \ENDWHILE
    \STATE $\myvec{x}_\text{new} = n_\text{nearest}$.state; $t_\text{extend}$ = 0
    \WHILE {not ($t_\text{extend} >$   $t_\text{maxExtend}$ or reach($\myvec{x}_\text{new}$, $\myvec{x}_\text{rnd}$) or $\myvec{x}_\text{new}$ is in collision)}  
        \STATE $t_\text{extend}$ += $\Delta t$
        \STATE $\myvec{o}$ = makeObservation($\myvec{x}_\text{new}$, $\myvec{x}_\text{rnd}$)
        \STATE $\myvec{x}_\text{new}$ = propagateDynamics($\myvec{\pi}(\myvec{o})$, $\myvec{x}_\text{new}$)
        \IF {$\myvec{x}_\text{new}$ is not in collision and $t_\text{extend}$ \% $\Delta t_\text{tree} = 0$}
            \STATE $\mathcal{T}$.add(makeNode($\myvec{x}_\text{new}$, $\myvec{x}_\text{rnd}$))
        \ENDIF
    \ENDWHILE
\ENDWHILE
\RETURN $\mathcal{P}$ = extractMotionPlan($\mathcal{T}$)

\end{algorithmic}
\caption{RL-RRT} 
\label{alg:rlrrt}
\end{algorithm}

Within \alg,  the obstacle-aware reachability estimator can provide insight into the best samples to enhance tree growth.  However, as we began to use the estimator, it became clear that the obstacle-aware reachability estimator can take longer than the standard Euclidean distance metric to compute (about 0.5 ms vs. 7 $\mu$s for Euclidean).  Therefore, to enhance computation time in large trees, the estimator was integrated into a hierarchical nearest neighbor selector.  Similar to  \cite{chiang2018fast}, the method first identifies $k_c$ candidate nodes closest to $\myvec{x}_\text{rnd}$ using Euclidean distance (Alg. \ref{alg:rlrrt}, line 8), and subsequently these choices are filtered by the obstacle-aware TTR between each candidate node and $\myvec{x}_\text{rnd}$.  To alleviate  noise in the TTR estimator, we take the average of the TTR between the selected node and $N_\text{TTR sample}$=10 target states around $\myvec{x}_\text{rnd}$, i.e., within a hypercube of $d_{TTR sample}$=0.3 units (line 10).
The node with the lowest average TTR is selected for 
RRT extension (line 9).
In addition, 
the obstacle-aware reachability estimator can also be used to check whether the randomly sampled state $\myvec{x}_\text{rnd}$ is reachable from the nearest node $n_\text{nearest}$.
Recall that the TTR reward in Section \ref{sec:methodReachability} is setup such that any $\myvec{x}_\text{rnd}$ unreachable from $n_\text{nearest}$.state has an associated V-value larger than $T_\text{horizon}$. 
As the result, the estimated TTR can be used to prune out $\myvec{x}_\text{rnd}$ that are un-reachable from the tree within $T_\text{horizon}$.
However, since the estimated TTR is not exact, we made the pruning probabilistic, i.e., if $\myvec{x}_\text{rnd}$ is deemed unreachable, it will be pruned with probability  $P_\text{prune}$ (line 10).
If $\myvec{x}_\text{rnd}$ is pruned, it is rejected and a new $\myvec{x}_\text{rnd}$ is sampled (line 6).

After the nearest node is selected, RL-RRT uses the RL policy $\myvec{\pi}$ as the local planner (lines 15-24). 
Specifically, an observation $\myvec{o}$ which includes \revised{3.8}{simulated lidar}, robot state, and goal information is made at every policy time step $\Delta t$ (line 17).
This observation is fed to the RL policy, which produces an action that can be used to forward propagate the dynamics to a new state $\myvec{x}_\text{new}$ (line 18).
This process repeats and a new node storing $\myvec{x}_\text{new}$ is created, and added to the tree every $\Delta_\text{tree}$ seconds (line 21), until $\myvec{x}_\text{new}$ is in collision, a maximum extension time is reached (line 20), or $\myvec{x}_\text{rnd}$ is reached (line 20).

RL-RRT terminates when either the tree reaches the goal or after a fixed amount of computation time is exhausted (line 3).
If the tree reaches the goal, a dynamically-feasible motion plan can be returned (line 25).


\section{Evaluation}
\label{sec:results}

\revised{3.7}{To demonstrate RL-RRT, we evaluate our method on three kinodynamic robots in two environments unseen during training}, and we experimentally verify the method on a physical differential drive Fetch robot from Fetch Robotics.

\subsection{Setup}
\label{sec:setup}

The three robots we evaluate are: Car, Asteroid, and Fetch. 
Car is a kinematic car with inertia \cite{paden2016survey} with a maximum steering angle $30^\circ,$ and a 1.0 $m/s^2$ maximum acceleration and speed of 1.0 $m/s$.
Asteroid has similar dynamics to those found in the popular video game Asteroid, and we chose it since it is highly kinodynamic, unintuitive for a human to control, and has no known optimal steering function. The details are available in the supplemental materials. 
The Fetch robot has a radius of \dist{0.3}, 1.0 m/s maximum speed and 2.0 rad/s turn rate. 
\revised{}{The sensor noise is simulated by a zero mean Gaussian with a standard deviation of 0.1 m.}
We use the Fetch robot as a differential drive platform for on-robot experiments.

All point-to-point policies are trained in the environment depicted in Figure \ref{fig:envTrain}.
We evaluate RL policies and plan in two office building environments, Map 1 (Figure \ref{fig:coverPic}) and Map 2 (Figure \ref{fig:physicalPath}), which are roughly 15 and 81 times larger than the training environment, respectively. 
Map 1 is is generated from a floor plan, while Map 2 is generated using a noisy SLAM of the Fetch physical testbed where we ran the experiments.
These environments include parts that are cluttered, as seen in Map 1, and very narrow corridors, such seen in Map 2.

We compare RL-RRT to SST \cite{li2016asymptotically}, a state of the art steering function free kinodynamic motion planner.
For Fetch robot, we also compare to RRT with Dynamic Window Approach (DWA) \cite{fox1997dynamic} as local planner (denoted RRT-DW). 
Additionally, we test disabling the clearance term of DWA, essentially turning it into a MPC-based steering function (denoted RRT-S). 
\revised{}{All experiment are repeated 50 times. Besides AutoRL training, all computation was done} on an Intel Xeon E5-1650 @ 3.6GHz using TensorFlow 1.x (Google release) and Python 2.7. AutoRL policies were implemented with Google Vizier \cite{vizier} and TFAgents \cite{tfagents}.

\subsection{AutoRL Policy Performance}
\label{sec:rlPerf}
We use pre-trained P2P policies for Fetch \cite{autorl} and Car \cite{francis2019long} robots.   Their short description is available in the Appendix. The Asteroid P2P policy is original to this paper. All agents are trained with AutoRL over DDPG \cite{autorl}. The goals are randomly placed within \dist{10}. 
\revised{}{We train 100 agents in parallel over 10 generations as in \cite{autorl}. The training took roughly 7 days.}

\begin{figure*}[h]
	\begin{center}
		\begin{tabular}{ccc}
			\subfloat[\scriptsize Differential Drive]{\includegraphics[width=0.3\textwidth,height=2.1cm,keepaspectratio=false]{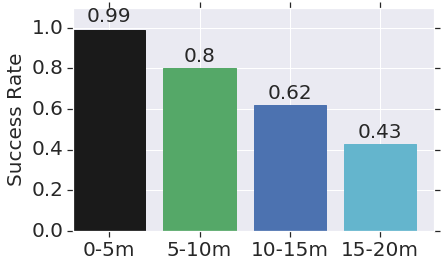}\label{fig:p2pSuccAst}}&
			
			\subfloat[\scriptsize Car]{\includegraphics[width=0.3\textwidth,height=2.1cm,keepaspectratio=false]{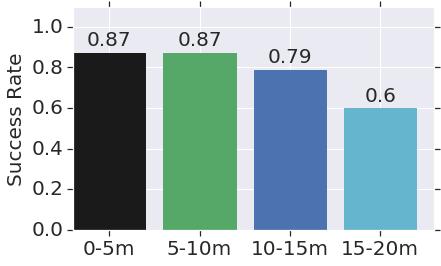}\label{fig:p2pSuccCar}}&
			
			\subfloat[\scriptsize Asteroid]{\includegraphics[width=0.3\textwidth,height=2.1cm,keepaspectratio=false]{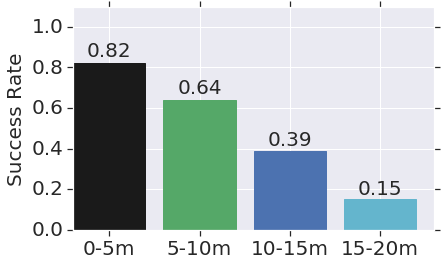}\label{fig:p2pSuccAsoh ot}}
			\\
		\end{tabular}
		\caption{\footnotesize AutoRL P2P navigation success rate as a function of start and goal distance for (a) Fetch, (b) Car and (c) Asteroid robot. \revised{3.6}{The success rates are evaluated in Map 1 with randomly sampled start and goal states.}
		\label{fig:p2pSucc}}
	\end{center}
\end{figure*}

Figure \ref{fig:p2pSucc} shows the success rate of the P2P agents compared to goal distance. Notice that when the goal distance is  \dist{10} or farther than the trained policy, the performance degrades. We also notice that the Car policy is best performing, while the Asteroid policy is the most challenging. 
These results show that AutoRL produces, without hand-tuning, effective local planners, i.e., both a steering function and an obstacle avoidance policy for a variety of robot dynamics. 

\subsection{Reachability Estimator Performance}
\label{sec:ttrEstimatorPerf}
The obstacle-aware reachability estimator is trained in the training environment with goals sampled within \dist{20} from the initial states, twice the distance used for P2P training. 
The estimator network was trained on 1000 episodes with about 100,000 samples. \revised{2.5}{Data generation takes about 10 minutes.}
The reachability thresholds are 20 seconds for differential drive and Asteroid, and 40 seconds for Car. 
\revised{2.5}{Each estimator was trained over 500 epochs and took about 30 minutes.}

\begin{table}[tb]
	\centering
	\begin{tabular}{l| l | r  r | r | r | r } 
         \multirow{ 2}{*}{Robot} & \multicolumn{3}{|c|}{Confusion Matrix} & Prec. & Recall &Accur. \\ \cline{2-4}
          & & \multicolumn{2}{|c|}{True (\%)} & (\%)  & (\%) & (\%) \\  \hline
        \multirow{ 2}{*}{Fetch} & \multirow{ 1}{*}{Predicted}    & 42.7 & 21.6 & \multirow{ 2}{*}{66.4} & \multirow{ 2}{*}{92.2} & \multirow{ 2}{*}{74.8} \\
        &  (\%) & 3.6  & 32.1 &  &  &  \\  \hline
        \multirow{ 2}{*}{Car} & \multirow{ 1}{*}{Predicted}    & 44.5 & 14.2 & \multirow{ 2}{*}{75.8} & \multirow{ 2}{*}{90.2} & \multirow{ 2}{*}{81.0} \\  
         &   (\%) & 4.8  & 36.5 &  &  &  \\ \hline
        \multirow{ 2}{*}{Asteroid} & \multirow{ 1}{*}{Predicted}    & 26.5 & 16.3 & \multirow{ 2}{*}{61.9} & \multirow{ 2}{*}{73.4} & \multirow{ 2}{*}{74.1} \\
        &   (\%) & 9.6  & 47.6 &  &  &  \\
        \hline
	\end{tabular}
	\caption{ \footnotesize
    Reachability estimator confusion matrix, precision, recall, and accuracy in the training environment.
	}
    \label{tab:reachConfusion}
\end{table}

Accuracy of the models is between 70\% and 80\% (Table \ref{tab:reachConfusion}). Notice that a high recall means that the estimator misses fewer nodes, and suggests that the paths RL-RRT produces should be near-optimal. On the other hand, relatively low precision implies that RL-RRT will explore samples that end up not being useful. This means that we can speed-up RL-RRT further by learning a more precise predictor.

\begin{figure*}[h]
\centering
\begin{tabular}{ccc}
			\subfloat[\scriptsize Differential Drive]{\includegraphics[width=0.3\textwidth,height=3.1cm,keepaspectratio=false]{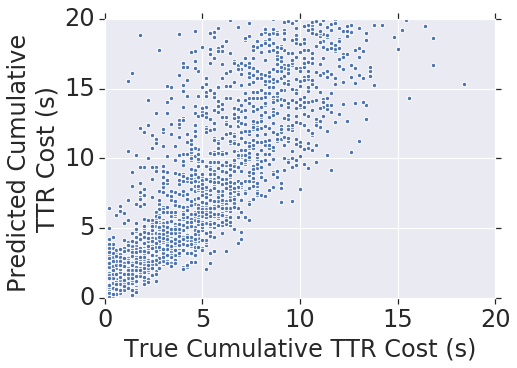}\label{fig:ttrScatterFetch}} &
			\subfloat[\scriptsize Car ]{\includegraphics[width=0.3\textwidth,height=3.1cm,keepaspectratio=false]{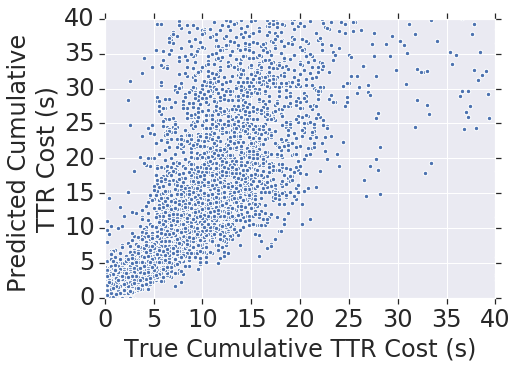}\label{fig:ttrScatterCar}}&
			
			\subfloat[\scriptsize Asteroid]{\includegraphics[width=0.3\textwidth,height=3.1cm,keepaspectratio=false]{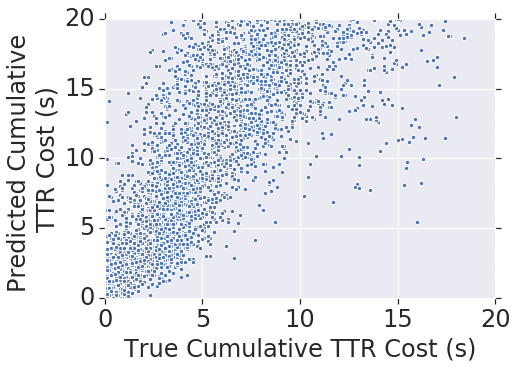}\label{fig:ttrScatterAst}}
		\end{tabular}
		\caption{\footnotesize \revised{3.11}{Predicted cumulative future time to reach cost v.s. true value for various robots.}}
		\label{fig:ttrScatter}
\end{figure*}

			

\revised{R3.2}{ The reachability estimator overestimates the TTR of reachable states across all robots (Fig. \ref{fig:ttrScatter}). However, overestimation disappears when trained and evaluated only on reachable states (see Fig. 1 in Appendix for more detail).
This suggests that the overestimation of TTR is likely due to the TTR cost heuristic uses a penalty for states unreachable within $T_\text{horizon}$.
We leave identifying better TTR cost heuristics and estimator network architectures for future work. 
}

\begin{figure*}[tb]
\centering
    \subfloat[\scriptsize Training environment (22.7 x 18.0 m)]{\includegraphics[width=0.2\textwidth,keepaspectratio=false,trim=40mm 0mm 40mm 0mm,clip]{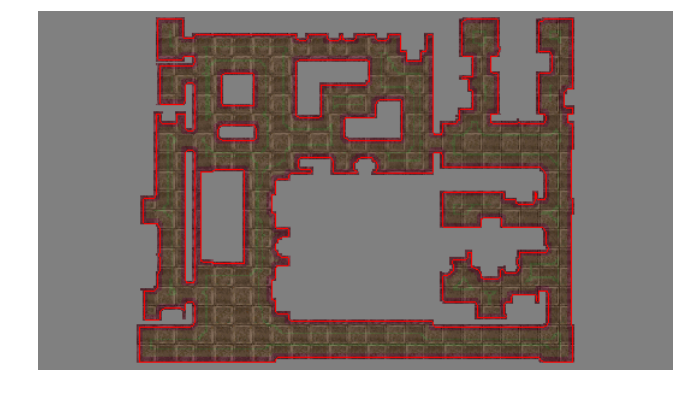}\label{fig:envTrain}}
	\subfloat[\scriptsize Predicted]{\includegraphics[width=0.3\textwidth,keepaspectratio=false]{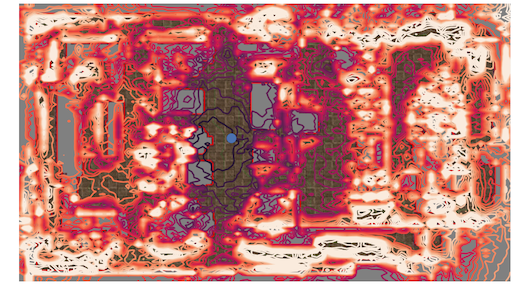}\label{fig:valueLandscapeNet}}
	\subfloat[\scriptsize Ground truth ]{\includegraphics[width=0.3\textwidth,keepaspectratio=false]{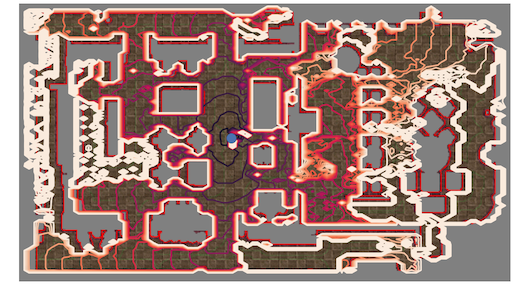}\label{fig:valueLandscapeTruth}}
	\caption{\footnotesize \revised{}{(a) The training environment. Contour plot of (b) Predicted \revised{3.6}{future cumulative time to reach cost} v.s. (c) the true value for \revised{3.6}{Car} to reach the goal near the center marked by the blue dot. The white regions have time to reach value over the 40s horizon, i.e., un-reachable. All start states and the goal have 0 as linear speed and orientation.}}
\label{fig:valueLandscape}		
\end{figure*}

In general, the estimator captures the regions of start states that cannot reach the goal (blue dot) (Fig. \ref{fig:valueLandscape}). 
This is most visible at the bottom right region of the environment, which has a TTR larger than the 40s horizon which indicates that the policy failed to escape that region.
We also see that the estimated TTR captures the dynamics of Car robot, i.e., since the goal orientation is facing right, it takes less time to reach the goal from the left, top or bottom than from the right.
Note that the network is never trained on trajectories that start inside of obstacles and thus cannot accurately predict TTR starting from those states, an event which should not occur in sampling-based planning.  


\subsection{Planning Results}
\label{sec:planningResults}

\begin{figure*}[h]
    \centering
    \addtocounter{subfigure}{-1}
    \subfloat{\includegraphics[width=0.6\textwidth,height=0.5cm,keepaspectratio=false]{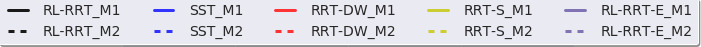}}
    
	\subfloat[\scriptsize Differential Drive.]{\includegraphics[width=0.3\textwidth,height=3.0cm,keepaspectratio=false]{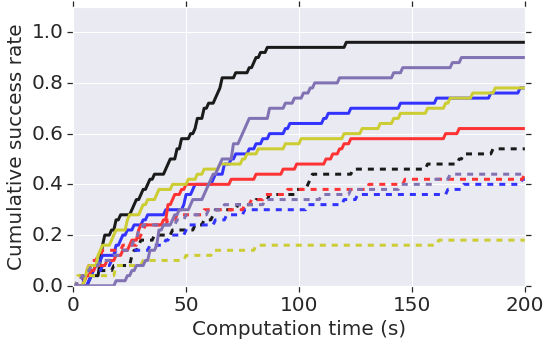}\label{fig:perfSuccFetch}}
	\subfloat[\scriptsize Car ]{\includegraphics[width=0.3\textwidth,height=3.0cm,keepaspectratio=false]{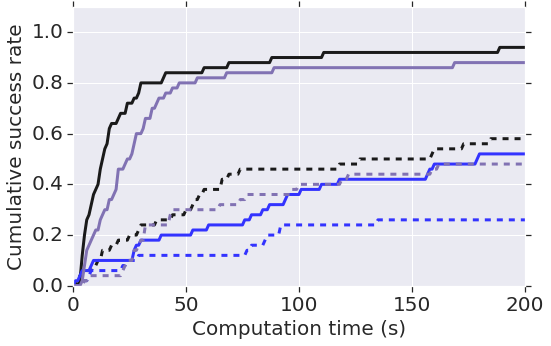}\label{fig:perfSuccCar}}
	\subfloat[\scriptsize Asteroid]{\includegraphics[width=0.3\textwidth,height=3.0cm,keepaspectratio=false]{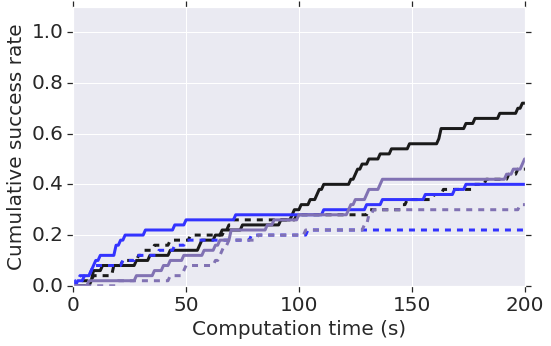}\label{fig:perfSuccAstroid}}
	
	\subfloat[\scriptsize Differential Drive.]{\includegraphics[width=0.3\textwidth,height=3.0cm,keepaspectratio=false]{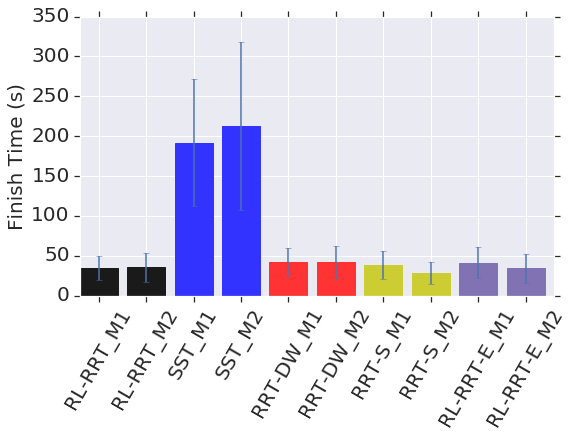}\label{fig:perfFinishFetch}}
	\subfloat[\scriptsize Car]{\includegraphics[width=0.3\textwidth,height=3.0cm,keepaspectratio=false]{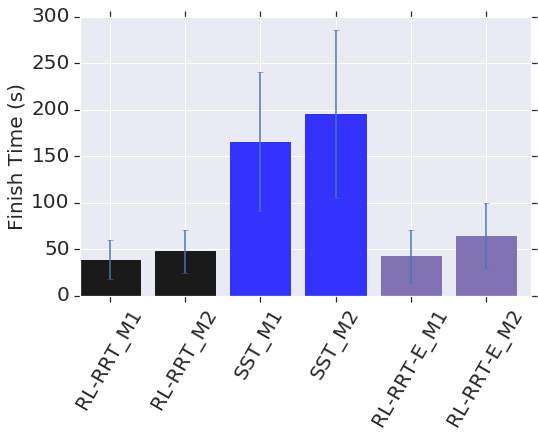}\label{fig:perfFinishCar}}
	\subfloat[\scriptsize Asteroid]{\includegraphics[width=0.3\textwidth,height=3.0cm,keepaspectratio=false]{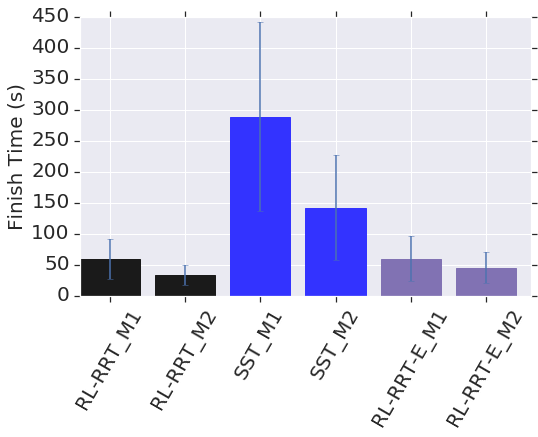}\label{fig:perfFinishAstroid}}
	
    \caption{\footnotesize Success rate (top) and Finish time (bottom) of RL-RRT (black) compared to, SST (blue), RRT-DW (red, RRT with DWA obstacle-avoiding steering function), RRT-S (yellow, RRT with DWA as the steering function) and RL-RRT-E (magenta, RL-RRT using Euclidean distance instead of the reachability estimator) in Map 1 (M1) and Map 2 (M2). }
    \label{fig:perf}
\end{figure*}

RL-RRT finds a solution faster than SST for all three robots in both environments (Fig. \ref{fig:perfSuccFetch}, \ref{fig:perfSuccCar}, \ref{fig:perfSuccAstroid}). Note that Car shows the best improvement over the baseline (up to 2.3 times faster), which matches the high success rate of the P2P Car policy. 
Conversely, the least improvement is for Asteroid, which is the most challenging for the RL agent. 
Figure \ref{fig:perfSuccFetch} also shows that RL-RRT finds a solution faster than steering function-based methods, where DWA was used as the steering function (yellow, RRT-S) and obstacle-avoiding steering function (red, RRT-DW).
These results are expected as RL-RRT learns a obstacle-avoiding local planner that can often go through very narrow corridors and move around corners (Figure \ref{fig:coverPic}).
In comparison, DWA often gets stuck around corners.
To separate the impact of the RL local planner as compared to the reachability estimator, we tested RL-RRT without the estimator and use Euclidean distance to identify the nearest state in the tree instead. 
Figures \ref{fig:perfSuccFetch}, \ref{fig:perfSuccCar} and \ref{fig:perfSuccAstroid} show that RL-RRT without the reachability estimator (magenta curves) performs worse than RL-RRT for all robots.
This is expected as the reachability estimator prunes potentially infeasible tree-growth, thereby biasing growth towards reachable regions. 
Also, the reachabilty estimator encodes the TTR and is thus more informative than the Euclidean distance for kinodynamic robots such as Asteroid.

The finish time of trajectories identified by RL-RRT are significantly shorter (up to 6 times shorter) than SST for all robots (Fig. \ref{fig:perfFinishFetch}, \ref{fig:perfFinishCar}, \ref{fig:perfFinishAstroid}) and comparable to RRT-DWA and RRT-S on differential drive.
This is expected as SST does not use steering functions. Instead, it randomly propagates actions, resulting in a ``jittery'' behavior (visible in Figure \ref{fig:coverPic}) and long finish time.
The comparable finish time with steering function-based methods show that RL-RRT learns a near-optimal steering function.


\subsection{Physical Robot Experiments}
\label{sec:physical}

In order to verify that the RL-RRT produces motion plans that can be used on real robots, we executed the motion plans on the Fetch robot (Figure. \ref{fig:fetchPic}) in Map 2 environment.
We ran 10 different motion plans, repeated 3 times. 
Figure \ref{fig:physicalPath} presents one such trajectory. The straight line distance between the start and goal is 20.8 m. In green are tree nodes for a path, and the blue line is the executed robot path with the P2P AutoRL policy. We notice two things. First, the path is similar to the one humans would take. The shortest path leads through cubicle space, which is cluttered. Because the P2P policy does not consistently navigate the cubicle space, the TTR estimates are high in that region and the tree progress slowly in that area. At the same time, in the uncluttered space  near the start position (left and right) the tree grows quickly. The executed trajectory (in blue) stays close to the planned path.  
Enclosed video contains the footage of the robot traversing the path. 

\section{DISCUSSION}
\label{sec:discussion}

\begin{figure}[h]
    \centering
	\subfloat[\scriptsize Two Astroid trajectories.]{\includegraphics[width=0.24\textwidth,height=2.4cm,keepaspectratio=false]{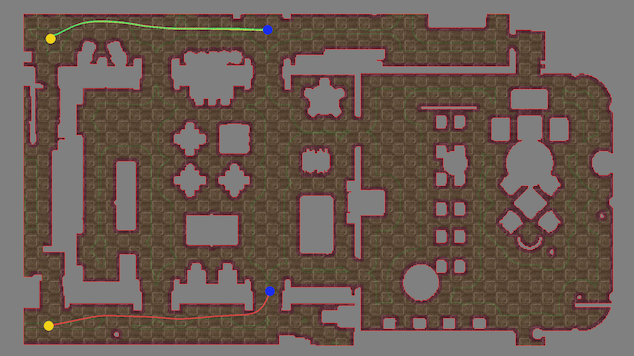}\label{fig:ddpgTraj}}
	\subfloat[\scriptsize V-value and TTR.]{\includegraphics[width=0.25\textwidth,height=2.4cm,keepaspectratio=false]{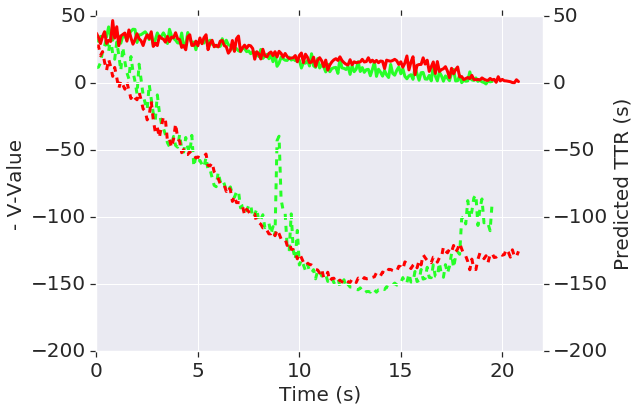}\label{fig:ddpgVals}}
	
    \caption{\footnotesize (a) Two trajectories (green and red) of the Asteroid robot from the yellow dots to blue dots. (b) The corresponding predicted TTR \revised{3.6}{(solid lines)} and the negative of V-value from DDPG's critic net \revised{3.6}{(dashed lines)}.}
    \label{fig:xrndReach}
\end{figure}

\revised{3.9}{Deep actor-critic RL methods approximate the cumulative future reward, i.e., state-value function with the critic net. Intuitively, the state-value function captures the progress towards the goal and may be used as a distance function during planning.
Here we show that this is \textit{not} the case when proxy rewards are used. AutoRL uses proxy rewards (shown in Section \ref{sec:methodAutoRl}) since they significantly improve learning performance, especially for tasks with sparse learning sigals such as navigation \cite{autorl}.}
Fig \ref{fig:ddpgTraj} shows examples of two Asteroid trajectories and Fig. \ref{fig:ddpgVals} shows the corresponding the estimated TTR (solid lines) and negative of DDPG state-value function extracted form the critic net (dashed lines). The obstacle-aware reachability estimator correctly predicted the TTR while the DDPG's critic net has a significant local maximum, thus unsuitable as a distance function. This finding motivated the supervised reachability estimator.

\begin{figure}[h]
    \centering
	\subfloat[\scriptsize Predicted]{\includegraphics[width=0.18\textwidth]{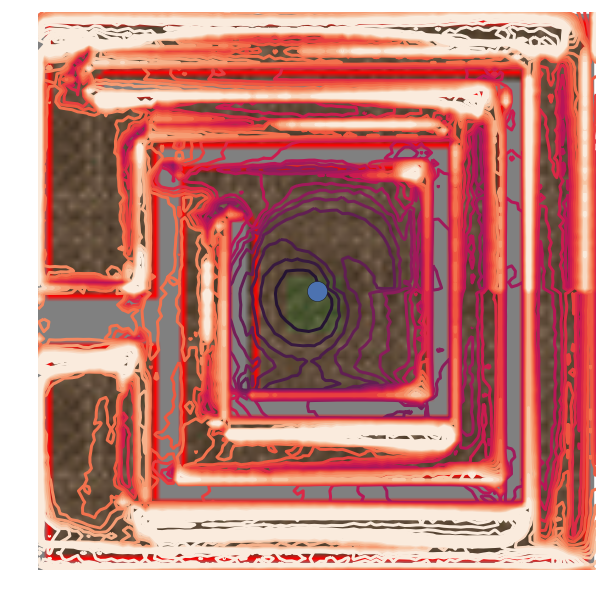}\label{fig:valueLandscapeNetMaze}}
	\subfloat[\scriptsize Ground truth ]{\includegraphics[width=0.18\textwidth]{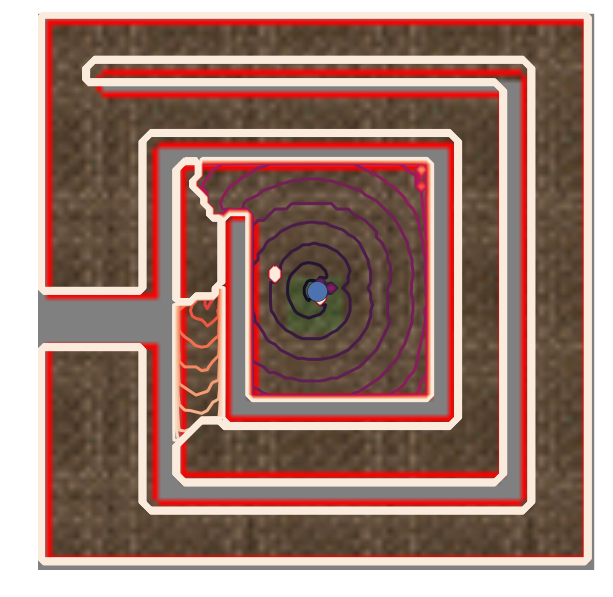}\label{fig:valueLandscapeTruthMaze}}
	
    \caption{\footnotesize \revised{1.1}{Contour plot of (a) Predicted future cumulative time to reach cost v.s. (b) the true value for Car to reach the goal near the center marked by the blue dot. The white regions have time to reach value over the 40s horizon, i.e., un-reachable. All start states and the goal have 0 as linear speed and orientation. The environment size is 50 m by 40 m.}}
    \label{fig:xrndReachMaze}
\end{figure}

\revised{1.1}{One limitation of RL-RRT is that the obstacle-aware reachability estimator approximates reachability using only local information such as simulated lidar measurements around the robot. 
However, the true reachability is often impacted significantly by large-scale obstacle structures.
Figure \ref{fig:xrndReachMaze} demonstrates this limitation. 
The ground truth shows that the Car policy generally fails to reach the goal outside of the center box due to the complex maze-like obstacles (Figure \ref{fig:valueLandscapeTruthMaze}). 
The reachability estimator failed to predict this as some regions outside of the center box are incorrectly predicted as reachable (Figure \ref{fig:valueLandscapeNetMaze}). 
On the other hand, we also demonstrated that the estimator performs well when the training and planning environments are similar (Figure \ref{fig:valueLandscape}).
This suggests that the reachability estimator should to be trained in environments similar to the planning environment or perform online adaptation/learning during planning. 
We leave the latter to future work.
}

	


\section{CONCLUSIONS}
\label{sec:conclusions}
This paper contributes RL-RRT, a kinodynamic planner which works in three steps: 1) learning obstacle-avoiding local planner; 2) training an obstacle-aware reachability estimator for the learned local planner; and 3) using the estimator as the distance function and to bias sampling in RRT. 
\revised{3.7}{Unlike traditional kinodynamic motion planners, RL-RRT learns a suitable steering and distance function. The robot is trained once, and the policy and estimator transfer to the new envrionments.}
We evaluated the method on three kinodynmic robots in two simulated environments. Compared to the baselines, RRT plans faster and produces shorter paths. We also verified RL-RRT on a physical differential drive robot.
\revisedV{For future work, following PRM-RL, we plan to improve the noise robustness of RL-RRT by Monte Carlo roll-outs during tree extensions.} We also plan to identify better TTR cost heuristics, network architectures and online adaptation of the reachability estimator.





\section*{ACKNOWLEDGMENT}
\small
We thank Tsang-Wei Edward Lee for assisting with robot experiments, and Brian Ichter for the helpful feedback.   
Tapia and Chiang are partially supported by the National Science Foundation under Grant Numbers IIS-1528047 and IIS-1553266 (Tapia, CAREER). 
Any opinions, findings, and conclusions or recommendations expressed in this material are those of the authors and do not necessarily reflect the views of the National Science Foundation.


\bibliographystyle{abbrv}

\bibliography{literature}

\section*{Supplemental Material: \\
Rewards for the P2P}

\subsection{P2P for differential drive robots}
\label{sec:p2p-diff-drive}
The P2P agent was developed in \cite{autorl}. 
The reward is:
\begin{equation}
R_{\myvec{\theta_{r_\text{DD}}}} = 
\myvec{\theta_{r_\text{DD}}}^T [
r_\text{goal}
r_\text{goalDist} \,
r_\text{collision} \,
r_\text{clearance} \,
r_\text{step} \,
r_\text{turning} \,
],
\label{eq:p2p_reward}
\end{equation}
where
$r_\text{goal}$ is 1 when the agent reaches the goal and 0 otherwise,
$r_\text{goalDist}$ is the negative Euclidean distance to the goal,
$r_\text{collision}$ is 1 when the agent collides with obstacles and 0 otherwise,
$r_\text{clearance}$ is the distance to the closest obstacle,
$r_\text{step}$ is a constant penalty step with value 1, and
$r_\text{turning}$ is the negative angular speed.

\subsection{P2P for kinodynamic car robots}
\label{sec:p2p-car-model}
The P2P agent was developed in \cite{francis2019long}. The robots dynamics is identical as  \revised{}{\cite{paden2016survey}}. The reward is:
\begin{equation}
R_{\myvec{\theta_{r_\text{CM}}}} = 
\myvec{\theta_{r_\text{CM}}}^T [
r_\text{goal}
r_\text{goalProg} \,
r_\text{collision} \,
r_\text{step} \,
r_\text{backward} \,
],
\label{eq:p2p_reward}
\end{equation}
where $r_\text{goal}$, $r_\text{collision}$ and $r_\text{step}$ are the same as the differential  drive.
$r_\text{goalProg}$ rewards the delta change of Euclidean distance to the goal.
$r_\text{backwards}$ is the negative of backwards speed and is zero when the robot moves forward.
$N_\text{beam}=64$.

\section*{Supplemental Material: \\ Asteroid}
Asteroid has a similar dynamics to those found in the popular video game Asteroid.
\begin{align}
    \ddot{x} = a_\text{thrust} cos(\theta) - \kappa \dot{x} \\
    \ddot{y} = a_\text{thrust} sin(\theta) - \kappa \dot{y} \\
    \dot{\theta} = a_\theta
\end{align}
$a_\text{thrust}$ is the thruster acceleration action ranged from [-0.5, 1.0] $m/s^2$ while $a_\theta$ is the turn rate action ranged from [-0.5, 0.5] rad/s. $\kappa=1.0$ $s^{-1}$ is the first order drag coefficient, resulting in a maximum speed of 1.0 $m/s$.
$N_\text{beam}=64$.

\section*{Supplemental Material: \\ Time To Reach Estimators}

\begin{figure}[h]
\centering
		\subfloat[\scriptsize Differential Drive]{\includegraphics[width=0.24\textwidth,keepaspectratio=false]{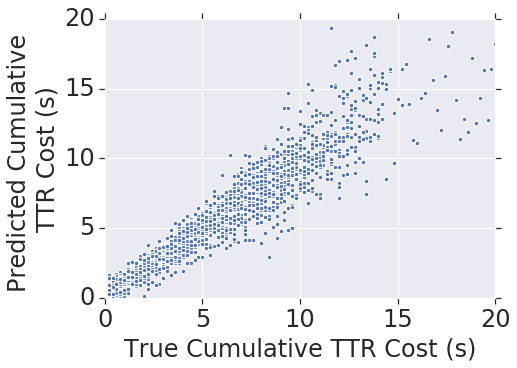}\label{fig:ttrScatterFetchSucc}}
		\subfloat[\scriptsize Car ]{\includegraphics[width=0.24\textwidth,keepaspectratio=false]{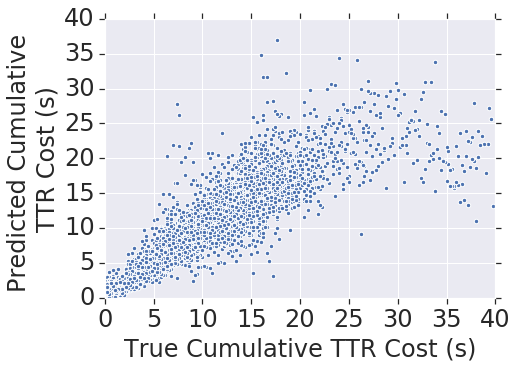}\label{fig:ttrScatterCarSucc}}
		
		\subfloat[\scriptsize Asteroid]{\includegraphics[width=0.24\textwidth,keepaspectratio=false]{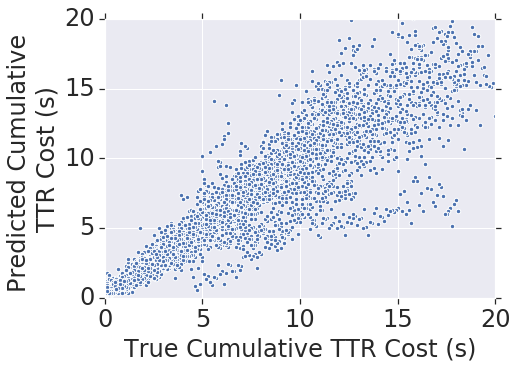}\label{fig:ttrScatterAstSucc}}
		\caption{\footnotesize \revised{3.11}{Predicted time to reach v.s. true value for various robots. The estimators are trained and evaluated with only states that can reach the goal.}}
		\label{fig:ttrScatterSuccOnly}
\end{figure}

\revised{3.2}{The obstacle-aware reachability estimator combines rechable state classification and TTR estimation in order to bias tree-growth towards reachable regions and identifying nearest neighbors. 
Here we explore estimating only the TTR by training a TTR estimator that is trained only by trajectories that reached the goal.
Fig. \ref{fig:ttrScatterSuccOnly} shows the predicted TTR and the ground truth for various robots. 
Unlike the reachability estimator (Fig. 4 in the main paper), the TTR estimator does not overestimate TTR. This suggests that the overestimation of the reachability estimator is caused by the TTR cost heuristic penalizing unreachable states.}

\end{document}